\documentclass[runningheads]{llncs}
\usepackage{xcolor}          % 推荐用 xcolor，更灵活
\usepackage[colorlinks=true,   % 用纯颜色而不是方框
            citecolor=black,    % 引用颜色（最常用）
            linkcolor=black,    % 内部链接（如 \ref）
            urlcolor=black,     % URL 颜色
            filecolor=magenta  % 本地文件链接颜色（可选）
]{hyperref}
\usepackage{amssymb} % add new pkg
\usepackage{bbding} % add new pkg
\usepackage[T1]{fontenc}
\usepackage{graphicx,verbatim}
\begin{document}
\title{EndoGSim: Physics-Aware 4D Dynamic Endoscopic Scene Simulations via MLLM-Guided Gaussian Splatting}
% \title{EndoGSim: Automatic Physical-Aware 4D Dynamic Endoscopic Simulations via MLLM-Guided Gaussian Splatting}
\begin{comment}  %% Removed for anonymized MICCAI submission
\author{First Author\inst{1}\orcidID{0000-1111-2222-3333} \and
Second Author\inst{2,3}\orcidID{1111-2222-3333-4444} \and
Third Author\inst{3}\orcidID{2222--3333-4444-5555}}
%
\authorrunning{F. Author et al.}
\institute{Princeton University, Princeton NJ 08544, USA \and
Springer Heidelberg, Tiergartenstr. 17, 69121 Heidelberg, Germany
\email{lncs@springer.com}\\
\url{http://www.springer.com/gp/computer-science/lncs} \and
ABC Institute, Rupert-Karls-University Heidelberg, Heidelberg, Germany\\
\email{\{abc,lncs\}@uni-heidelberg.de}}

\end{comment}

% \author{Anonymized Authors}  %% Added for anonymized MICCAI submission
\author{Changjing Liu\inst{1}\thanks{Co-first authors.} \and
Yiming Huang\inst{1}\protect\footnotemark[1] \and
Long Bai\inst{1} \and
Beilei Cui\inst{1} \and
Hongliang Ren\inst{1}\thanks{Corresponding author.}}
\authorrunning{Anonymized Author et al.}
\institute{Department of Electronic Engineering, The Chinese University of Hong Kong (CUHK), Hong Kong SAR, China\\
\email{\{changjingliu,yhuangdl\}@link.cuhk.edu.hk, b.long@ieee.org, beileicui@link.cuhk.edu.hk, hlren@ee.cuhk.edu.hk}}
  
\maketitle              % typeset the header of the contribution
\begin{abstract}
In robot-assisted minimally invasive surgery, high-fidelity dynamic endoscopic scene reconstruction and simulation are crucial to enhancing downstream tasks and advancing surgical outcomes.
However, existing methods primarily focus on visual reconstruction, lacking 
physics-based descriptions of the scene required for realistic simulation.
We propose a unified framework that achieves physics-aware reconstruction and physical simulation of endoscopic scenes through Multi-modal Large Language Models (MLLMs)-guided Gaussian Splatting.
Our approach utilizes 4D Gaussian Splatting (4DGS) integrated with pre-trained segmentation and depth estimation to represent deformable tissues and tools.
To achieve automatic inference of physical properties, we introduce an object-wise material field that initializes material parameters via MLLM and refines them through a differentiable Material Point Method (MPM) under joint supervision from rendered images and optical flow.
Validated on both open-source and in-house datasets, our framework achieves superior simulation fidelity and physical accuracy compared to state-of-the-art methods, underscoring its potential to advance robot-assisted surgical applications.
% Our approach has been validated on both opens-source and in-house datasets, where it can realize reconstruction and simulation with remarkable physical accuracy and realism.
% render in real-time, compute efficiently, and reconstruct with remarkable accuracy.
% This integrated framework enables accurate prediction and realistic simulation of dynamic interactions in real-world scenarios, advancing both accuracy and flexibility in physics-based simulations. 

% Authors must provide keywords and are not allowed to remove this Keyword section.

\end{abstract}

\section{Introduction}
Endoscopic procedures are central to minimally invasive surgery with reduced trauma and faster recovery, motivating the integration of robot-assisted systems to achieve superior precision, consistency, and efficiency~\cite{raptis2026artificial,ding2026telerobotic,chen2025robotic}.
% Endoscopic procedures have become a cornerstone of minimally invasive surgery, offering reduced trauma and faster patient recovery.
% Consequently, automation in endoscopic robotic systems is critical for enhancing procedural precision, consistency, and efficiency~\cite{raptis2026artificial,ding2026telerobotic,chen2025robotic}.
% In light of these advancements, automation in endoscopic robotic systems has emerged as a critical direction for enhancing procedural precision, consistency, and efficiency~\cite{raptis2026artificial,ding2026telerobotic,chen2025robotic}.
% Surgical automation in robotic systems is becoming the cornerstone in minimally invasive surgery, providing patients with less trauma and faster recovery, and enhance procedural consistency.
% In this context, accurate and dynamic spatial understanding of the endoscopic scene—achieved via precise 3D reconstruction and realistic dynamic simulation—is critical to enhancing the surgeon’s navigation, facilitating more precise and efficient interventions, and enabling surgical automation in robotic systems.
% In this context, accurate dynamic reconstruction and simulation of the endoscopic scene are essential for reliable spatial intelligence and robust autonomous control.
In this context, accurate dynamic reconstruction and simulation of the endoscopic scene are essential to enable reliable spatial intelligence, support pre-operative planning, and ensure robust autonomous control~\cite{huang2024endo}.
Recent endoscopic scene reconstruction benefits from Neural Radiance Fields (NeRFs)~\cite{mildenhall2021nerf,wang2022neural,zha2023endosurf} and 3D Gaussian Splatting (3DGS)~\cite{kerbl20233d,yang2024deformable,chen2024mvsplat}, achieving superior reconstruction quality along with real-time rendering performance.
% Recent endoscopic reconstruction benefits from Neural Radiance Fields (NeRFs)~\cite{mildenhall2021nerf,wang2022neural,zha2023endosurf} for notable 3D vision, further enhanced by 3D Gaussian Splatting (3DGS)~\cite{zhu2024endogs} for faster training and real-time rendering via explicit differentiable primitives.
Building upon this foundation, 4D Gaussian Splatting (4DGS)~\cite{wu20244d,huang2024endo,gao2025endord} extends static representations to the spatiotemporal domain, making it suitable for deformable surgical scenes reconstruction.

\begin{table*}[h]
\caption{Comparison of methods by Video input, Auto Initialization, Object-wise material field, and Physical simulation.}
\label{tab:comparison}
\centering
\small
\begin{tabular}{l|c|c|c|c|c|c}
\hline
{Method}              &{PhysGen} & {Physics3D} & {PhysGaussian} & \phantom{xx}{GIC}\phantom{xx} & {PhysFlow} & \textbf{Ours} \\
\hline
Video Input
& \XSolidBrush 
& \XSolidBrush
& \XSolidBrush
& \Checkmark
& \Checkmark
& \Checkmark \\
Auto Initialization
& \Checkmark
& \XSolidBrush
& \XSolidBrush
& \XSolidBrush
& \Checkmark
& \Checkmark \\
Material Field
& \XSolidBrush
& \XSolidBrush
& \XSolidBrush
& \XSolidBrush
& \XSolidBrush
& \Checkmark \\
Physical Simulation
& \XSolidBrush
& \Checkmark
& \Checkmark
& \Checkmark
& \Checkmark
& \Checkmark \\
\hline
\end{tabular}
\end{table*}

Despite remarkable advances in scene reconstruction, existing methods~\cite{kerbl20233d,wu20244d} primarily focus on appearance reconstruction and seldom capture the material properties of objects.
% , resulting in incomplete scene reconstruction~\cite{xie2024physgaussian}.
This limits physically accurate spatial perception, high-fidelity dynamic simulation, and the generation of realistic data essential for surgical automation.
To address this challenge, several studies integrate material property estimation and perform simulations on reconstructed deformable scenes~\cite{xie2024physgaussian,cai2024gic,liu2024physics3d,liu2024physflow}.
% PhysGaussian~\cite{xie2024physgaussian} first combines 3DGS with a differentiable Material Point Method (MPM) to enhance simulation fidelity, but it relies on manual material property initialization, limiting adaptability to complex dynamic scenes.
PhysGaussian~\cite{xie2024physgaussian} combines 3DGS with a differentiable Material Point Method (MPM) for physics-based simulation, with material parameters specified through manual initialization.
% 需要改
% GIC~\cite{cai2024gic} employs 3DGS to capture explicit shapes and leverages continuum mechanics to infer physical property. 
% However, it requires multi-view images and prior shape knowledge, restricting its applicability to general dynamic videos.
To enable automatic identification of material properties, GIC~\cite{cai2024gic} captures explicit shapes using 3D Gaussian Splatting and infers physical properties through continuum mechanics for simulation.
% GIC~\cite{cai2024gic} captures explicit shapes with 3D Gaussian Splatting (3DGS) and infers physical properties via continuum mechanics. 
% This method relies on multi-view images and prior shape knowledge, limiting its applicability to general dynamic monocular videos.
% Recent methods~\cite{liu2024physics3d,liu2024physflow} employ video diffusion models to guide motion and estimate material properties.
% , but often fail to deliver physically accurate realism.
% However, SDS-based video generation frequently fails to achieve true physical realism.
% In contrast, we investigate that image and optical flow guidance is both memory-efficient and better suited for capturing large and complex motions, enabling more accurate material parameter optimization.
% More recent works~\cite{liu2024physgen,liu2024physflow} incorporate Multi-modal Large Language Models (MLLMs) to enable automatic material reasoning and enhanced dynamic simulation, with PhysGen leveraging physics-aware reasoning and Phyflow exploiting multimodal priors and video diffusion.
Subsequent works~\cite{liu2024physgen,liu2024physflow} further integrate multi-modal large language models (MLLMs) to enhance physics-aware reconstruction and dynamic simulation.
% By leveraging the prior knowledge encoded in MLLMs through visual and textual inputs, material parameters can be initialized within physically plausible ranges.
PhysGen~\cite{liu2024physgen} introduces physics reasoning using large pre-trained visual foundation models, eliminating the need for manual parameter initialization. 
Physflow~\cite{liu2024physflow} leverages multi-modal foundation models and video diffusion to achieve enhanced 4D dynamic scene simulation.

\begin{figure*}[t]
  \begin{center}
    \centerline{\includegraphics[width=0.95\columnwidth]{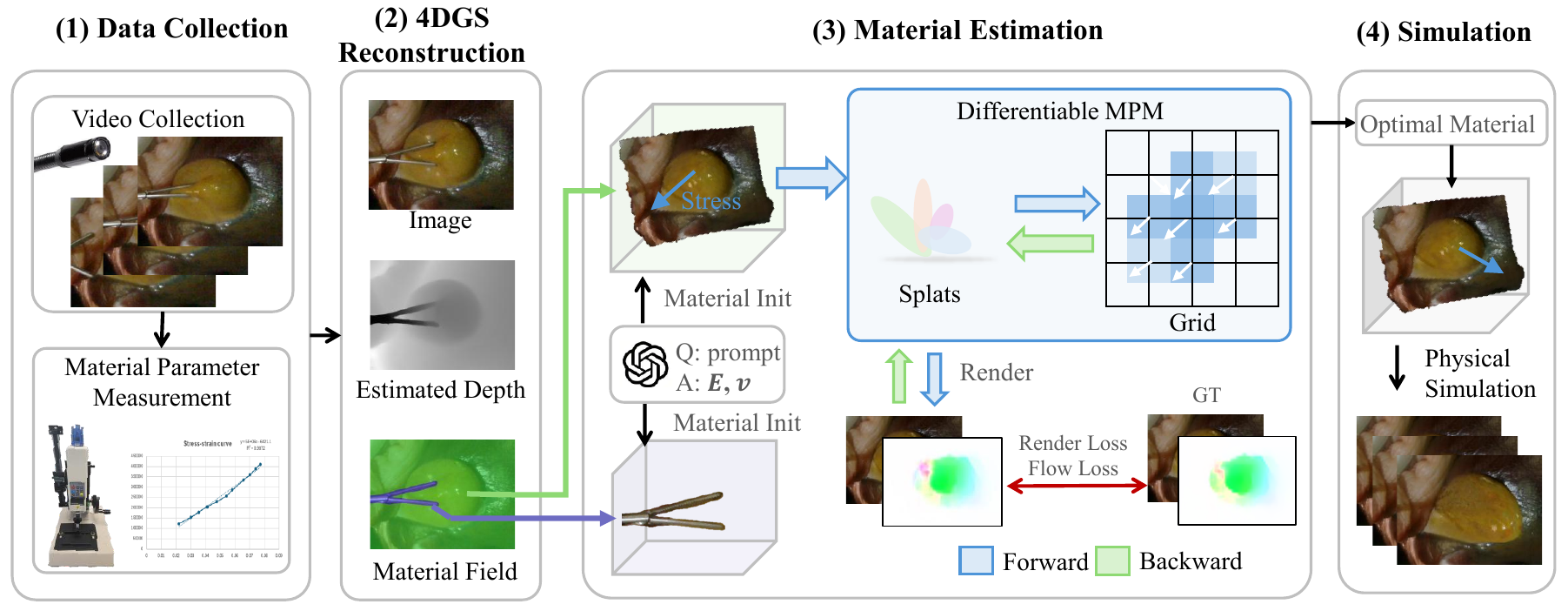}}
    \caption{
    Overview of our physics-aware framework for surgical scene reconstruction and 4D dynamic simulation with automatic estimation of physical parameters.
      % Overview of our physics-aware framework for surgical scene reconstruction and 4D dynamic simulation with automatic physical parameter estimation.
      % The process starts with data collection and material measurement, followed by scene reconstruction via 4D Gaussian Splatting
      % Initial material properties of reconstructed scene are inferred through multi-modal foundation models.
      % The material parameters are optimized using optical flow and render guidance, integrated with a differentiable MPM to ensure physically realistic simulation.
    }
    \label{fig:overview}
  \end{center}
\end{figure*}

However, there remains a notable gap in adapting these techniques for physics-aware reconstruction and simulation of surgical scenes.
In particular, 3DGS-based physical simulation methods~\cite{xie2024physgaussian,liu2024physics3d} struggle with dynamic video input, and existing material estimation strategies~\cite{xie2024physgaussian,cai2024gic,liu2024physics3d} often lack informed initialization, potentially leading to physically inconsistent simulations.
Furthermore, existing approaches mainly focus on general scenes and rarely consider multi-object material estimation, which is critical in surgical environments involving both instruments and deformable tissues.
To overcome these challenges, we propose a unified framework that reconstructs deformable endoscopic scenes from video and simultaneously estimates material parameters to enable realistic physical simulation, as shown in Fig.~\ref{fig:overview}.
% To enable accurate reconstruction from endoscopic video, we employ 4DGS~\cite{huang2024endo} with $\pi^3$ depth estimation~\cite{wang2025pi} and Surgical-SAM-2~\cite{liu2024surgical} segmentation to generate the Gaussian splats representation for the surgical scene. 
Specifically, we employ 4DGS~\cite{huang2024endo} with pre-trained depth and segmentation models to construct a Gaussian splat representation of the surgical scene.
Then, we propose an object-wise material field to estimate the physical properties of the tissues and tools.
Material parameters are automatically initialized using large pre-trained multi-modal large language models, and then refined via gradient-based optimization within a differentiable MPM, guided by rendering and optical flow, to ensure both visual realism and physical fidelity.
% Specifically, our method used large pre-trained visual foundation models (e.g., GPT-4~\cite{achiam2023gpt}) to initialize object material parameters for physically grounded simulations.
% To enable automatic refinement, we propose using rendering and optical flow as guidance within a differentiable MPM, allowing gradient-based optimization of material parameters to match observed interactions and achieve both visual plausibility and physical fidelity in the simulations.
Table~\ref{tab:comparison} summarizes the comparison of our approach with existing methods. 
Code and datasets will be released upon publication. 
The primary contributions of this work are summarized as follows:
% 重新写
\begin{itemize}
% \item A unified framework was introduced to reconstruct deformable scenes from endoscopic videos while estimating material parameters for physical simulation.
% \item We introduce a framework for reconstructing deformable scenes from endoscopic videos while simultaneously estimating material parameters to enable realistic physical simulation.
\item We introduce a unified framework for the automatic, physics-aware reconstruction and simulation of surgical scenes from endoscopic videos.
% We leverage a MLLM to infer initial material properties, while jointly reconstructing 3D Gaussian splats with a material field to support multi-object dynamic scenes.
\item The material field is proposed for material estimation, coarsely initialized via MLLM estimation and then jointly refined with render and optical flow loss in a differentiable MPM, achieving physically plausible simulation.
% \item Experiments on open-source and in-house datasets demonstrate that our method achieves superior physical consistency and enables more realistic simulation, offering substantial potential to accelerate surgical automation.
\item Experiments on open-source and in-house datasets demonstrate that our method achieves accurate physics and realistic simulations, offering substantial potential to advance robotic-assisted minimally invasive surgery.
\end{itemize}

\section{Methods}
\subsection{Preliminaries}
% Our approach adopts the representation scheme proposed in PhysGaussian~\cite{xie2024physgaussian}, where the 3D geometry is discretized into a set of unstructured Gaussian primitives $\{x_p, \sigma_{p}, A_p, C_p\}_{p{\in}P}$. Specifically, $x_p$, $\sigma_p$, $A_p$ and $C_p$ denotes the centers, opacities, covariance matrices, and spherical harmonic coefficients of the Gaussians, respectively. During the rendering phase, these 3D kernels are transformed into 2D screen-space distributions. The final pixel color is obtained by integrating the contributions of sorted Gaussians along the camera ray using the volume rendering equation:
% \begin{equation}
%     C = \sum_{k{\in}P} \alpha SH(d_K;C_k)\prod_{j=1}^{k-1}(1-\alpha_j).
% \end{equation}
% $\alpha_k$ represents the $z$-depth ordered effective opacities; $d_k$ stands fot the view direction from the camera to $x_k$.
Inspired by~\cite{huang2024endo}, we reconstruct the surgical scene via 4D Gaussian $\mathcal{G}'=\mathcal{G} + \Delta\mathcal{G}$, including a static 3D Gaussian $  \mathcal{G}  $ and its deformation $  \Delta\mathcal{G}=\mathcal{F}(\mathcal{G},t)  $, where $  \mathcal{F}  $ denotes the deformation network and $  t  $ is the time.
The spatial-temporal encoder $\mathcal{H}$ is defined to parameterize the 4D trajectory of each Gaussian point.

A multi-head Gaussian deformation decoder $\mathcal{D}$ is designed for decoding the deformation of position $\mu$, rotation $\textbf{R}$, scaling $\textbf{S}$, opacity $o$ and spherical harmonics $\textbf{SH}$ using five efficient and lightweight tiny MLPs. 
The final 4D Gaussian representation can be formulated as follows:
% The final representation of 4D Gaussian can be expressed as:
\begin{equation}
    \mathcal{G}'=\{\mu+ \Delta{\mu} ,\textbf{R} + \Delta{\textbf{R}}, \textbf{S} + \Delta{\textbf{S}}, o + \Delta{o},\textbf{SH} + \Delta{\textbf{SH}}\}.
\end{equation}

For dynamic scene video $V=\{I_0,...I_t\}$, we adopt 4D Gaussian Splatting to construct a time-dependent 3D scene representation.
Specifically, the $\pi^3$ model~\cite{wang2025pi} is integrated to provide depth estimation, and Surgical-SAM-2~\cite{liu2024surgical} is utilized for efficient segmentation of soft tissues and surgical instruments.
% These high-quality depth priors and segmentation masks provide crucial constraints for the 4D Gaussian optimization, enabling high-fidelity subsequence simulation.

\subsection{Continuum Mechanics}
Continuum mechanics provides the foundational description of the motion of soft tissues during surgical procedures.
% Continuum mechanics provides the foundational description of soft-tissue motion and deformation in surgical interventions.
Specifically, the deformation of a continuum is described via the time-dependent continuous map $\mathbf{x} = {\phi}(\mathbf{X}, t)$, where $\mathbf{X}$ represents the undeformed material space and $\mathbf{x}$ denotes the deformed space at time $t$. 
The deformation gradient $\mathbf{F}(\mathbf{X}, t)=\nabla _{\mathbf{X}}\phi(\mathbf{X}, t)$ captures the local deformation of the material, encoding its rotation, stretch, and shear.

% Our work employs a hyperelastic-viscoplastic model parameterized by a vector $  \theta_p = \{E, \nu, \tau_Y, \eta\}  $, where $  E  $ is Young’s modulus, $  \nu  $ is Poisson’s ratio, $  \tau_Y  $ is the yield stress, and $  \eta  $ is the plastic viscosity.
Our work employs a hyperelastic model parameterized by a vector $  \theta_p = \{E, \nu \}  $, where $  E  $ is Young’s modulus and $  \nu  $ is Poisson’s ratio.
% Our work models soft tissue behavior using a hyperelastic–viscoplastic model, with material parameters $\theta_p = \{E, \nu \}$ (Young’s modulus and Poisson’s ratio) optimized while the remaining coefficients are fixed to predefined values.
% In particular, we optimize the parameters Young’s modulus $E$ and Poisson’s ratio $\nu$ with a vector$\theta_p = \{E, \mu \}$, while the remaining viscoplastic coefficients are fixed to predefined values.
% Our work employs a hyperelastic model with material parameter  Young’s modulus $E$ and Poisson’s ratio $\nu$.
% Our work employs a hyperelastic-viscoplastic model characterized by Young’s modulus $E$, Poisson’s ratio $\nu$, yield stress $\tau_Y$, and plastic viscosity $\eta$.
To accurately represent these properties, the evolution of $ \phi  $ must satisfy mass and momentum conservation: mass conservation ensures uniform density during deformation, while momentum conservation is formulated as follows:
\begin{equation}
\label{Continuum Mechanics}
\rho(\mathbf{x}, t)\dot{\mathbf{v}}(\mathbf{x},t) = \nabla\cdot\mathbf{\sigma}(\mathbf{x},t) + \mathbf{f}^{ext},  
\end{equation}
where $\rho$ is the mass density, $\sigma$ is the Cauchy stress tensor, and $\mathbf{f}^{ext}$ denotes external forces (tool-tissue interactions).
The deformation gradient $  \mathbf{F} = \mathbf{F}^E \mathbf{F}^P  $ decomposes into elastic ($  \mathbf{F}^E  $) and plastic ($  \mathbf{F}^P  $) contributions, allowing realistic modeling of tissue elastic recoil and permanent tearing under surgical forces.
% Here the total deformation gradient $\mathbf{F} = \mathbf{F}^E \mathbf{F}^P$ can be decomposed into an elastic part $\mathbf{F}^E$ and plastic part $\mathbf{F}^P$. This decomposition is essential for simulating complex tissue behaviors, from elastic recoil upon unloading to permanent tearing under strong surgical force.
% To account for both reversible strain and permanent damage—such as the tearing of soft tissues—the deformation gradient $\mathbf{F}$ is multiplicatively decomposed into elastic and plastic components, $\mathbf{F} = \mathbf{F}^E \mathbf{F}^P$.

\subsection{Material Point Method (MPM)}
% The Material Point Method (MPM) \cite{stomakhin2013material} offers an efficient way to solve the continuum equations defined in Eq. \ref{Continuum Mechanics}.
% Different from mesh-based numerical mechanics, MPM can be naturally applied to point-based representation 3DGS following by PhysGaussian~\cite{xie2024physgaussian}.
The Material Point Method (MPM) \cite{stomakhin2013material,de2020material} provides an efficient solver for the continuum equations in Eq.~\ref{Continuum Mechanics} and naturally extends to point-based representations such as 3DGS, as demonstrated in PhysGaussian~\cite{xie2024physgaussian}.
% The \textbf{Differentiable Material Point Method } extends the classical Material Point Method (MPM) to enable gradient-based optimization and learning in continuum mechanics simulations. 
% In MPM, the continuum is represented by a set of Lagrangian particles $\{\mathbf{x}_p, \mathbf{v}_p, m_p, \mathbf{F}_p\}$ that carry the material state, while a background Eulerian grid is used to solve the equations of motion.
The MPM operates in a particle-to-grid (P2G), grid update, and grid-to-particle (G2P) transfer loop.
In P2G process, MPM transfers mass and momentum from particles to grids as:
\begin{equation}
    m^{n}_{i} = \sum_{p} w_{ip}^{n}m_p, 
    \quad 
    m^n_i\mathbf{{v}_{i}^{n}} = \sum_{p}w^{n}_{ip}m_p(\mathbf{{v}^{n}_{p}}+C_p^n(\mathbf{x_i}-\mathbf{{x}^n_p})),
\end{equation}
where $p$ and $i$ denote the fields on the Lagrangian particles and the Eulerian grid, respectively.
Each particle $p$ carries a set of physical properties, including its volume $V_p$, mass $m_p$, position $\mathbf{x}_p^n$, velocity $\mathbf{v}_p^n$, deformation gradient $\mathbf{F}_p^n$, and affine momentum ${C}_p^n$ at time $t_n$. The term $w_{ip}^n$ represents the B-spline kernel function centered at the $i$-th grid node and evaluated at the particle position $\mathbf{x}_p^n$.

After the P2G transfer, the velocities on the grid are updated by integrating both internal and external forces:
\begin{equation}
\mathbf{v}_i^{n+1} = \mathbf{v}_i^n - \frac{\Delta t}{m_i} \sum_p \tau_p^n \nabla w_{ip}^n V_p^0 + \Delta t \frac{\mathbf{f}^{ext}}{m_i},
\end{equation}
where $\tau_p^n$ is the Cauchy stress tensor, $\nabla w_{ip}^n$ is the gradient of the kernel function. In G2P, the velocities are transferred back to particles to update their states:
% the velocities are transferred back to the particles to update particle states:
\begin{equation}
   \mathbf{v}_p^{n+1} = \sum_i w_{ip} \mathbf{v}_i^{n+1}.
   % \quad \mathbf{x}_p^{n+1} = \mathbf{x}_p^n + \Delta t \mathbf{v}_p^{n+1}.
\end{equation} 

% The deformation gradient $\mathbf{F}_p$ is evolved as $\mathbf{F}_p^{n+1} = (\mathbf{I} + \Delta t \nabla \mathbf{v}_i) \mathbf{F}_p^n$ to track the history of strain and facilitate the simulation of complex elastoplastic material behaviors.
The deformation gradient $\mathbf{F}_p$ is updated to track changes, with adjustments made for plasticity as needed.
Specifically, our framework builds upon differentiable MLS-MPM~\cite{hu2018moving}, which delivers significantly improved forward simulation stability and efficient back-propagation capabilities.

\subsection{Object-wise Material Field Parameter Estimation}
% Object-wise material field parameter estimation involves two stages: initialization and optimization. 
% Specifically, we first initialize the material parameters for different objects using a foundation model, and then further refine them through a differentiable MPM pipeline, where the loss is computed and backpropagated to update the parameters.

\noindent \textbf{Material Field Initialization.}
Accurate material properties are essential for physically consistent deformations, and their initial values significantly influence the optimization process.
% To simulate accurately, knowing the material properties is essential for applying deformations that adhere to physical laws. 
% The initial values of material parameters play a significant role in the subsequent optimization process. 
% Therefore, utilizing foundation models to estimate initial material properties is critical for achieving realistic simulations. 
% 首先基于Surgical-sam-2对视频中不同物体进行分割，实现软体和刚体的区分。
% Then, we infer initial material properties by querying GPT-4 with the input image and questions ~\cite{achiam2023gpt}.
% In our pipeline, we first employ Surgical-SAM-2~\cite{liu2024surgical} to segment different objects within the video, enabling a clear distinction between soft tissues and rigid instruments $M_{tissue}, M_{tools} = SAM(I)$. 
In our pipeline, we first apply Surgical-SAM-2~\cite{liu2024surgical} to segment video frame $I$, obtaining distinct masks for tissues and instruments
$  M_{{tissue}}, M_{{tools}} = {SAM}(I)  $.
Subsequently, we infer initial material attributes $\theta_p^{init}$ by prompting GPT-4o~\cite{achiam2023gpt} with the segmented visual inputs: $\theta_p^{init} \leftarrow {GPT} ( {prompt}, I\odot M_{{tissue}}, \, I\odot M_{{tools}})$. The prompts are detailed in the supplementary material.
% For rigid bodies, these initialized parameters are sufficiently accurate and remain fixed. In contrast, for deformable objects, the parameters undergo further refinement to capture complex dynamics.
\\
\\
\noindent \textbf{Material Field Optimization.}
Optimization is selectively performed based on the object:
For rigid instruments and objects that undergo negligible deformation, the initialized parameters are frozen.
For deformable objects, we further optimize these parameters to better match the observed dynamics.
Specifically, given the input video sequence $V = \{I_0, \dots, I_t\}$, where $I_t$ denotes the observed frame at the time step $t$, we employ a pre-trained RAFT optical flow estimator~\cite{teed2020raft} to compute optical flow fields $U(I_t, I_{t+1})$ for the observed frames and $\hat{U}(\hat{I}_t, \hat{I}_{t+1})$ for the rendered frames $\hat{I}_t$. 
% We measure appearance differences between observed and rendered frames, and motion differences between the optical flow fields of observed and rendered consecutive pairs.
% We then compare the rendered images $\hat{I}_t$ with the observed frames $I_t$ and the corresponding flow fields across consecutive pairs, thereby quantifying both appearance discrepancies (via pixel-wise differences $\|I_t - \hat{I}_t\|_2^2$) and motion discrepancies (via flow differences $\|U(I_t, I_{t+1}) - \hat{U}(\hat{I}_t, \hat{I}_{t+1})\|_2^2$) between the simulation and observations. 
% The optimization process begins by analyzing the predicted motion from the simulation and comparing it to the
% optical flow-derived motion in the input frames.
% Then, we minimize a combined loss function $\mathcal{L}$ that leverages both appearance and motion supervision to align the simulation output $\hat{V}$ with the observed video $V$:
% We supervise both appearance (via per-pixel differences between $  I_t  $ and $  \hat{I}_t  $) and motion (via differences between corresponding optical flow fields). These cues are combined in the following joint loss:
% We supervise appearance (render loss between $  I_t  $ and $  \hat{I}_t  $) and motion (optical flow differences), combined via the joint loss:
We supervise appearance (rendering loss between $  I_t  $ and $  \hat{I}_t  $) and motion (optical flow differences) using a joint loss:
\begin{equation}
    \mathcal{L} = \sum_t \left\| I_t - \hat{I}_t \right\|_2^2 + \lambda \sum_t \left\| U(I_t, I_{t+1}) - \hat{U}(\hat{I}_t, \hat{I}_{t+1}) \right\|_2^2,
\end{equation}
% where $I_t$ and $\hat{I}_t$ denote the input and rendered frames at time $t$, respectively, 
% where $\hat{U}$ represent the optical flows computed from the observed frames, and $\lambda_{{flow}}$ balances the contributions of the two terms. 
% By integrating image supervision with flow-based guidance, our method more effectively captures transient object dynamics, leading to more accurate material property estimation than using render or optical loss alone.
% This information is then used to update the material parameters $M_p$, enabling the optimized simulation to capture the realistic observed behavior.
% This information is then used to update the material parameters via gradient descent $ \theta_p \leftarrow \theta_p - \eta \nabla_{\theta_p} \mathcal{L}  $ for physical simulation.
where $\lambda$ is set to 0.1 to balance the contributions of the loss terms.
The loss is then used to update the material parameters $\theta_p$ via gradient descent, enhancing the fidelity of the physical simulation.
% By integrating rendering supervision with optical flow guidance, our method effectively captures both appearance and motion variations, resulting in more accurate material property estimation than methods relying on a single supervisory signal.
%By combining both rendering and optical flow guidance, our method captures both appearance and motion more effectively, resulting in superior physical property estimation for simulation.

% After optimizing the material parameters $M_p$, the updated properties are integrated into the simulation to control the motion and deformation of the Gaussian splats ${G_p}$ with the differentiable MPM framework.

% To model physical properties, we employ MLS-MPM ~\cite{hu2018moving} as our simulator, and we formalize the
% simulation process for a single sub-step as follow: 
% \begin{equation}
%     \mathbf{x}^{n+1}, \mathbf{v}^{n+1}, \mathbf{F}^{n+1}, 
% \end{equation}

\section{Experiments}

\begin{table}[h]
\centering
\caption{ System identification performance (RE $\downarrow$ / EPE$\downarrow$) on the all dataset.}
\label{tab:system-performance}
\begin{tabular}{l|c|c|c|c|c}
\hline
{Method} 
&{PhysG.}~\cite{xie2024physgaussian} 
&{Physics3D}~\cite{liu2024physics3d} 
&{GIC}~\cite{cai2024gic} 
&{PhysFlow}~\cite{liu2024physflow} 
& \textbf{Ours} \\
\hline
v01\_080  
& 0.150/0.059  
& 0.230/0.071  
& 0.096/0.066 
& 3.836/6.368 
& \textbf{0.081}/\textbf{0.051} \\
v01\_240  
& 0.617/0.408
& 0.834/0.336 
& 0.129/0.068
& 3.290/1.601
& \textbf{0.035}/\textbf{0.016} \\
pulling
& 1.672/1.650
& 0.491/1.014 
& 1.048/1.769
& 4.651/1.683 
& \textbf{0.260}/\textbf{0.166} \\
cutting
& 2.041/0.592
& 0.251/0.141 
& 0.235/0.136 
& 0.196/0.142
& \textbf{0.146}/\textbf{0.077} \\
gallbladder 
& 0.192/0.617 
& 3.417/7.740
& 0.380/0.700
& 0.296/0.609
& \textbf{0.165}/\textbf{0.293} \\
stomach 
& 0.187/0.612 
& 0.964/1.578
& 0.199/\textbf{0.461}
& 0.668/0.988 
& \textbf{0.147}/0.501 \\
\hline
Avg.
& 0.810/0.657 
& 1.031/1.803
& 0.348/0.533
& 2.157/1.898 
& \textbf{0.139}/\textbf{0.184} \\
\hline
\end{tabular}
\end{table}

\begin{table}[ht]
\centering
\caption{Physical realism scores by surgeons, rated on a 5-point Likert scale (1 = strongly disagree to 5 = strongly agree) per video.}
\label{tab:Physical Realism Scores}
\begin{tabular}{l|c|c|c|c|c}
% \hline
%  & \multicolumn{5}{c|}{\textbf{Physical Realism (Score $\uparrow$)}} \\
\hline
{Method} 
&{PhysG.}~\cite{xie2024physgaussian} 
&{Physics3D}~\cite{liu2024physics3d} 
&{GIC}~\cite{cai2024gic} 
&{PhyFlow}~\cite{liu2024physflow} 
& \textbf{Ours} \\
% \hline
% Medical Experts & 3.89 & 3.69 & 3.78 & 3.52 & \textbf{3.98} \\
% General Public     & 3.43 & 3.07 & 3.76 & 3.07 & \textbf{4.05} \\
\hline
Physical Realism $\uparrow$        & 3.63 & 3.33 & 3.77 & 3.26 & \textbf{4.02} \\
\hline
\end{tabular}
\end{table}
\subsection{Data Curation}
Our dataset comprises two parts: 1. Two open-source surgical datasets. 2. An in-house dataset.
% \subsubsection{Open-source Benchmarks}:
We incorporate sequences from public EndoNeRF~\cite{wang2022neural} and CholecSeg8K~\cite{hong2020cholecseg8k} following~\cite{wang2022neural,huang2025surgtpgs}, while assigning ground-truth material parameters to the simulated scenes.
%, covering diverse surgical interactions to evaluate the robustness of our method.
% \subsubsection{In-house Benchmarks}:
Our in-house dataset PorcineEndo was constructed through an ex-vivo experiment involving fresh porcine stomach and gallbladder tissues.
% Video sequences were recorded using a monocular endoscope, ensuring high-fidelity visual capture for tissue reconstruction. 
% Besides, we conducted uniaxial tensile tests on tissue specimens using a HANDPI HLD Digital Force Gauge. 
Video sequences were recorded using a monocular endoscope, and uniaxial tensile tests were conducted on tissue specimens using a HANDPI HLD digital force gauge.
% The Young’s modulus was determined by applying a linear least-squares regression to the elastic region of the stress-strain curves. 
% For each specimen, 30 data points were sampled, yielding a high fitting accuracy with an average $R^2 > 0.99$.
The Young’s modulus was computed via a 30-point linear least-squares regression within the elastic region for each specimens, achieving accuracy with an average $R^2 > 0.99$.
Additionally, the Poisson's ratio was set to incompressible in this in-house dataset.
% The data curation pipeline is shown in Fig. ~\ref{fig:overview}. 
% This process ensures that our dataset contains both rich appearance information and accurate biomechanical properties, enabling the validation of our object-wise material field estimation.

% 自己的数据怎木采集的
% In this section, we showcase our method’s ability to automatically optimize material parameters and simulations, and evaluate its qualitative and quantitative effectiveness across diverse endoscopic datasets.
\subsection{Implementation Details}
All experiments are conducted on the A6000 GPU with the Python PyTorch framework.
% In our implementation, 4DGS reconstruction parameters follow~\cite{huang2024endo}, and both $\pi^3$ and Surgical-SAM-2 are employed with default settings.
% In our implementation, the 4D Gaussian Splatting reconstruction parameters follow in~\cite{huang2024endo}, while both $  \pi^3  $ and Surgical-SAM-2 are employed using their default settings.
Reconstruction is performed using 4DGS with parameters from~\cite{huang2024endo}, while pre-trained $\pi^3$~\cite{wang2025pi} and Surgical-SAM-2~\cite{wang2025pi} are used with default settings.
% We follow the 4D Gaussian Splatting reconstruction parameters from~\cite{huang2024endo}, and both pre-trained $\pi^3$~\cite{wang2025pi} and Surgical-SAM-2~\cite{wang2025pi} are employed with their default settings.
% The Gaussian kernels are then associated with physical properties for optimization.
The simulation is based on the warp~\cite{macklin2022warp} implementation of MLS-MPM.
% The simulation region is discretized onto a structured grid, typically with a resolution of $50^3$.
The simulation region is discretized onto a structured grid with a typical resolution of $  50^3  $ to enable efficient particle-grid interactions in the framework.
% For MLS-MPM~\cite{macklin2022warp} simulation, 
Specifically, we employ 25 sub-steps between successive renderings and include the gradient computed at the last sub-step in the optimization.
The sub-step duration is set to $2\times10^{-4}$, ensuring precision and accuracy in the simulation.
% On our setup, the whole material estimation process requires approximately 15 minutes per scene.
% For each scene, the simulation duration is $2\times10^{-4}$ s and frame duration $5\times10^{-3}$ s.
% Thus, we simulate 25 sub-steps between renderings and include the simulation gradient of the last step in the optimization.
\\
% \textbf{datasets}
% 数据集包含没有gt的数据集和有gt真实数据，没有gt的数据集以仿真的材料参数作为gt；有gt的数据集以真实的材料参数作为gt
% Our real-world evaluation includes a comprehensive set of scenes to cover diverse endo environment.
% endoNerf,steremis,endovis18.
% To enhance dataset diversity and coverage, we incorporate two additional self-collected scenes, including stomach.
% \textbf{Baseline}
% For real-world simulation, we compare our
% method with the following baselines:
% \\
% \textbf{Metric.}
% (1)human voting
% (2)FID,FVD
% (3)Energy-Constrained Motion Score (ECMS)

\begin{figure}[t]
  \vskip 0.2in
  \begin{center}
    \centerline{\includegraphics[width=0.85\columnwidth]{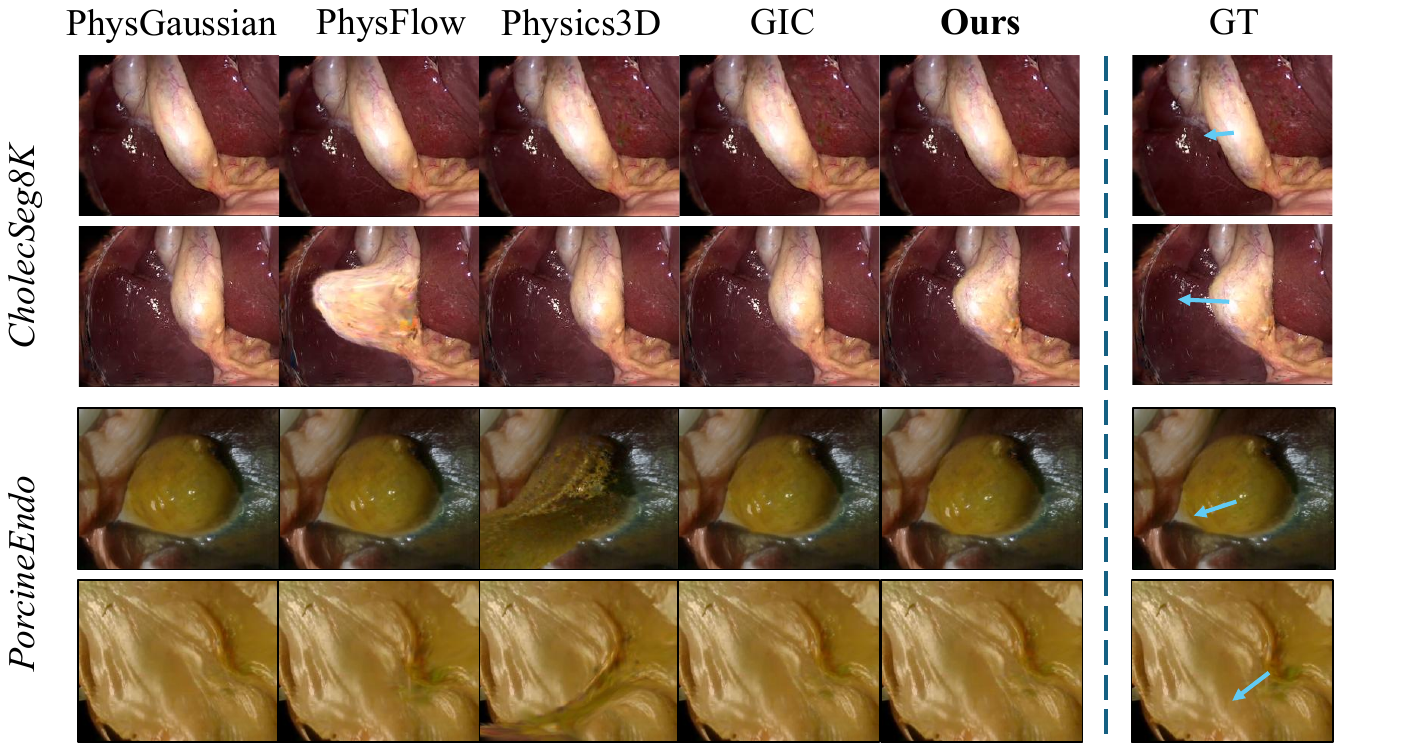}}
    \caption{
    % Qualitative results on EndoNeRF and PorcineEndo datasets. Results are shown for representative frames extracted from the simulation sequences.
    Qualitative results on the EndoNeRF and PorcineEndo datasets, shown for representative frames from the simulation sequences.
    }
    \label{fig:qualitative_result}
  \end{center}
\end{figure}

\subsection{Results}
% \textbf{Dataset} We use open-source endoscopic datasets, including EndoNerf~\cite{wang2024endogslam}(\textit{pulling} and \textit{cutting}), CholecSeg8K~\cite{hong2020cholecseg8k}(\textit{video01\_00080} and \textit{cutting\_tissues}).
% Since the open-source dataset lacks prior material properties, we assign specified material parameters and boundary conditions in simulation to generate ground-truth supervisory videos. 
% We additionally collect our own endoscopic dataset, including \textit{liver}, \textit{stomach}, and \textit{gallbladder}. 
% Our work focuses on validation of our effectiveness in optimizing material parameters, we adopt the 3D Gaussian splats object generated from the first frame of dynamic reconstruction by Endo-4dgs~\cite{huang2024endo} as our simulation objects.

\begin{figure}[t]
  \vskip 0.2in
  \begin{center}
    \centerline{\includegraphics[width=0.85\columnwidth]{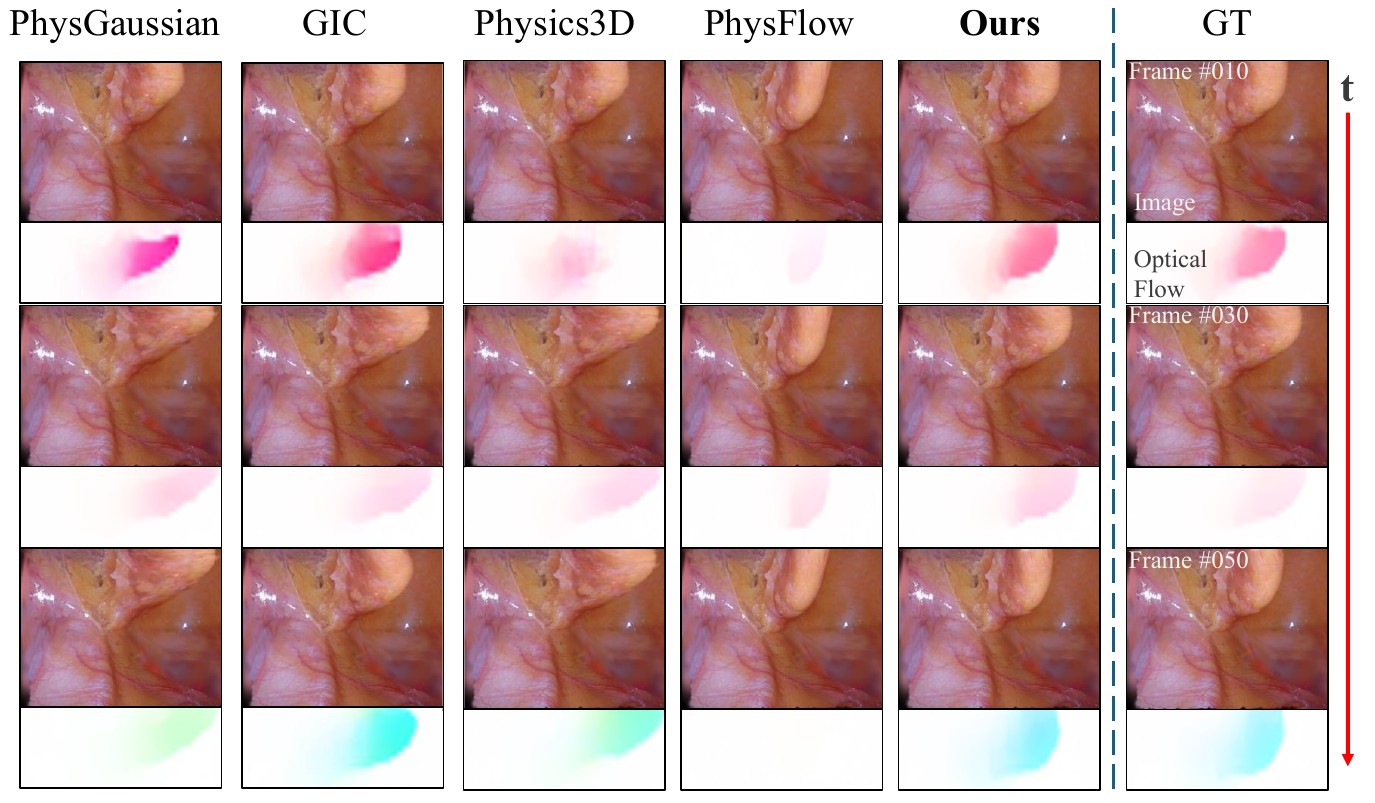}}
    \caption{
      % Qualitative simulation results of all methods on a sequence from the EndoNeRF dataset, illustrating rendered images and optical flow errors.
      Qualitative comparison of simulation results from all methods on a sequence of the EndoNeRF dataset, illustrating rendered images and optical flow errors.
      }
    \label{fig:flow_result}
  \end{center}
\end{figure}

% \textbf{Baseline} 
% We conducted a comprehensive comparison of our proposed method with state-of-the-art approaches (PhysGaussian(PhysG.))~\cite{xie2024physgaussian}, GIC~\cite{cai2024gic}, Physics3D~\cite{liu2024physics3d}, and PhysFlow~\cite{liu2024physflow}) for surgical scene reconstruction and simulation. 
We conducted a qualitative and quantitative comparison of our proposed method with state-of-the-art approaches, including PhysGaussian (PhysG.)~\cite{xie2024physgaussian}, GIC~\cite{cai2024gic}, Physics3D~\cite{liu2024physics3d}, and PhysFlow~\cite{liu2024physflow}, for surgical scene reconstruction and simulation. 
% Specifically, for a fair comparison under a unified supervision setting, We adapted the baselines for comparison: GIC is trained with rendering loss, while Physics3D and PhysFlow use ground-truth for guidance directly, bypassing their default video generation stage.
Specifically, for a fair comparison under a unified supervision setting, GIC is trained with rendering loss, whereas Physics3D and PhysFlow are trained with direct image supervision instead of generative images.
All baselines and our method optimize parameters $\theta_p$ starting from the same boundary settings per sence, such as including grid resolution, external force.
We evaluate performance using Relative Error (RE)~\cite{dagli2025vomp} for the estimated material parameters (Young’s modulus $E$, Poisson’s ratio $\nu$, shear modulus $G$, and bulk modulus $K$; $E$, $G$, and $K$ in log-space), End-Point Error (EPE)~\cite{dosovitskiy2015flownet} for optical flow accuracy, and Fréchet Inception Distance (FID)~\cite{heusel2017gans} for simulation realism.
As shown in Tab.~\ref{tab:system-performance}, our method consistently achieves lower RE and EPE than Physics3D and PhysFlow in most cases and remains competitive with GIC, demonstrating its effectiveness in estimating material properties and capturing deformation. 
% Our method exhibits a slightly higher EPE than GIC on the stomach scene, primarily because the relatively uniform color and large deformation in this scenario reduce the reliability of local optical flow estimation.
% Additionally, expert ratings from 33 surgeons indicate that our method obtains the highest realism score (Tab.~\ref{tab:Physical Realism Scores}), highlighting its potential for clinical applications such as haptic feedback training.
Additionally, expert ratings from 33 surgeons indicate that our method obtains the highest realism score (Tab.~\ref{tab:Physical Realism Scores}), highlighting its potential for clinical applications such as haptic feedback training.
% Additionally, Expert ratings from 33 surgeons show our method obtaining the highest realism score,  (Tab.~\ref{tab:Physical Realism Scores}), indicating its substantial potential for clinical training applications, such as haptic feedback training.
The qualitative results in Figs. \ref{fig:qualitative_result} and \ref{fig:flow_result} demonstrate that our approach generates more realistic and stable simulations with superior capture of material deformation than the baselines.

\begin{table}[t]
\centering
\caption{Ablation results (RE $\downarrow$ / EPE$\downarrow$ / FID$\downarrow$) on the PorcineEndo dataset. The best results are in bold.}
\label{tab:ablation_MLLM}
\begin{tabular*}{\textwidth}{@{\extracolsep{\fill}} c| c c| c c | c c | c c @{}}
\hline
{Initialization} & \multicolumn{2}{c|}{Manual} & \multicolumn{2}{c|}{Gemini-3-pro~\cite{gemini3pro}} & \multicolumn{2}{c|}{Claude-sonnet-4.5~\cite{claude4.5}} & \multicolumn{2}{c}{GPT-4o}~\cite{achiam2023gpt} \\
% \cline{2-9} 表示横线从第2列画到第9列，从而避开第1列的 "Optim." 标签空间
% \cline{2-9}
% \cline{2-3} \cline{4-5} \cline{6-7} \cline{8-9}
\hline
{Optimization} & \XSolidBrush & \Checkmark & \XSolidBrush & \Checkmark & \XSolidBrush & \Checkmark & \XSolidBrush & \Checkmark \\
\hline
RE $\downarrow$ & 0.309 & 0.160 & 0.197 & 0.064 & 0.193 & 0.103 & 0.144 & \textbf{0.063} \\
EPE $\downarrow$ & 0.938 & 0.514 & 0.701 & 0.414 & 0.679 & 0.417 & 0.604 & \textbf{0.397} \\
FID $\downarrow$ & 32.28 & 22.69  & 22.78 & 13.43 & 17.68 & 13.81  & 18.27 & \textbf{13.35}  \\
\hline
\end{tabular*}
\end{table}
% \subsection{Ablation Study}
% \subsubsection{Effectiveness of foundation model inference }
% \vspace{-10pt}
% To evaluate the impact of using a foundation model for physics reasoning, we conducted ablation experiments cross-comparing manual versus model-inferred material properties, both with and without subsequent optimization.
% We further conduct an ablation study that compares initialization methods (manual vs. MLLM) for material properties, with and without subsequent optimization.
We perform ablation experiments on material property initialization (manual vs. MLLMs), both with and without subsequent optimization.
% As shown in Table \ref{tab:ablation_MLLM}, manual initialization produces motions that only partially capture realistic behavior with the lowest performance, and subsequent optimization of these parameters yields moderate realism gains.
As shown in Tab.~\ref{tab:ablation_MLLM}, manual initialization yields partial realism and the lowest performance, and subsequent optimization brings moderate improvement.
% In contrast, MLLM-inferred values without optimization provide a markedly better starting point.
% By initializing with MLLM predictions and optimizing, our method achieves more accurate material consistency and visually convincing simulations.
MLLM-inferred properties provide a better starting point without optimization, and MLLM initialization followed by optimization delivers the highest material consistency and visual realism.
We further ablate the object-wise material field to evaluate its effectiveness.
% As shown in Fig. \ref{fig:ablation2}, simulations without the material field predict higher stiffness for tissue, resulting in unrealistic resistance to deformation.
% With the material field, the MLLM assigns object-wise parameters, enabling more accurate material estimation and realistic simulations.
As shown in Fig. \ref{fig:ablation2}, the material field allows object-wise parameter assignment, resulting in more accurate material estimation and physically realistic simulations.
The ablation results demonstrate that each component contributes to improved accuracy, physical realism, and overall performance.

% Incorporating the material field allows the MLLM to assign object-specific parameters (e.g., lower Young's modulus for soft tissues, higher for rigid tools), enabling more precise material differentiation and simulations closer to real-world physics.
% As shown in Fig. \ref{fig:ablation2}, simulations without a material field predict excessively high soft-body stiffness, decreasing performance metrics.
% By incorporating the material field, the MLLM can interpret object-wise physical parameters (e.g., lower Young's modulus for soft tissues, higher for rigid tools).
% This enables more precise differentiation of material properties, resulting in simulations that more closely align with real-world physical behavior.

% \vspace{-5pt}
\begin{figure*}[t]
  \centering
  % \vspace{0.2in}
  \label{fig:ablation2}
  % \hspace{0.1em}  % 替换 \hfill 为小间距（0.8em ~ 0.5cm 左右）
  \begin{minipage}{0.25\textwidth}
    \centering
    \small
    \begin{tabular}{l|r|r}
      \hline
      {Metric} & w/o MF & w/ MF \\
      \hline
      RE  $\downarrow$ & 0.433 & \textbf{0.178} \\
      EPE $\downarrow$  & 0.366 & \textbf{0.132} \\
      FID $\downarrow$  & 8.81 & \textbf{4.01} \\
      \hline
    \end{tabular}
  \end{minipage}
  \hfill   % ← 关键！这个让两个 minipage 被推到两侧，实现并排
  \begin{minipage}{0.62\textwidth}
    \centering
    % \hspace*{1em}
    \includegraphics[width=\linewidth]{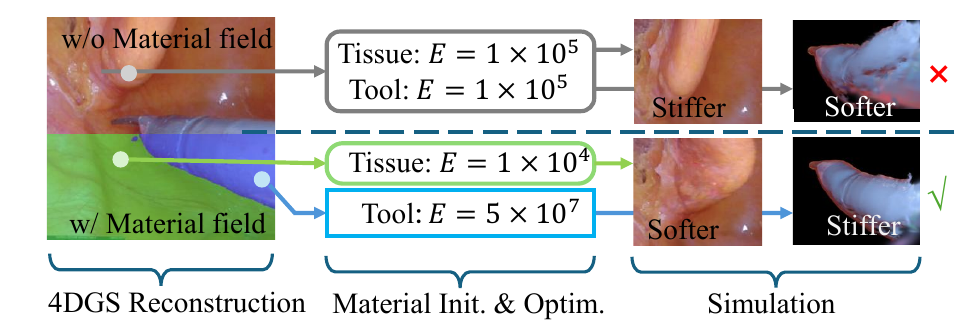}
  \end{minipage}%

  \caption{
    % (Left) Ablation results on the EndoNerf and CholecSeg8K dataset, where MF stands for Material Field.
    % (Right) Qualitative comparison: with Material Field vs. without
    Ablation on Material Field (MF): quantitative results on EndoNeRF and CholecSeg8K datasets (left), and qualitative comparison with vs. without MF (right).
    % Ablation on the Material Field (MF): results on the EndoNeRF and CholecSeg8K datasets (left), and qualitative comparison with and without MF (right).
  }
  \label{fig:ablation2}
\end{figure*}

% \begin{figure}[ht]
%   \vskip 0.2in
%   \begin{center}
%     \centerline{\includegraphics[width=0.98\columnwidth]{image/ablation_material_field.png}}
%     \caption{
%       Ablations on physics reasoning, showing material values, timestep 50 frame, and deformation.
%     }
%     \label{fig:ablation2}
%   \end{center}
% \end{figure}

% \begin{table}[h]
% \centering
% \caption{Ablation results (RE * $\downarrow$ / EPE $\downarrow$ / FID $\downarrow$) on the synthetic dataset.}
% \label{tab:Ablation results2}
% \small
% \begin{tabular}{|l|r|r|}
% \hline
% Metric            & w/o MF & w/ MF          \\
% \hline
% RE * $\downarrow$ & 0.063  & \textbf{0.040} \\
% EPE $\downarrow$  & 0.440  & \textbf{0.329} \\
% FID $\downarrow$  & 11.28  & \textbf{13.10} \\
% \hline
% \end{tabular}
% \end{table}

% \begin{table}
% \centering
% \caption{ Ablation results (RE * $\downarrow$ / EPE $\downarrow$ / FID $\downarrow$)  on the synthetic dataset.}
% \label{tab:Ablation results2}
% \begin{tabular}{|l|l|l|l|}
% \hline
% Varient &  RE * $\downarrow$ & EPE $\downarrow$ & FID $\downarrow$\\
% \hline
% \textbf{w/o MF}  & 0.063 & 0.440 &  11.28 \\
% \textbf{w/ MF} & \textbf{0.040} & \textbf{0.329} & \textbf{13.10}\\
% \hline
% \end{tabular}
% \end{table}

\section{Conclusion}
In this paper, we propose a physics-aware framework for reconstruction and simulation of surgical scenes from endoscopic videos.
We first integrate 4DGS with pre-trained depth estimation and segmentation to generate the Gaussian splats representation for the surgical scene.
The material field is proposed for material estimation, coarsely initialized via MLLMs-guide estimation and then jointly refined with render and optical flow loss in a differentiable MLS-MPM.
% Our method focuses on system identification of material parameters from visual observations, where optical flow is exploited to guide the optimization process. 
The optimized material properties are then incorporated into the simulation pipeline to enable realistic modeling of dynamic surgical scenes.
% Extensive experiments on open source and in-house datasets demonstrate that the proposed approach consistently outperforms existing baselines in terms of accuracy and realism, highlighting its potential for surgical scene simulation.
Qualitative and quantitative experiments on open-source and in-house datasets demonstrate that the proposed approach consistently outperforms existing baselines on accuracy and realism, highlighting its potential to advance robotic-assisted minimally invasive surgery with comprehensive perception capabilities.
\bibliographystyle{splncs04}
\bibliography{main}
% %
% % \begin{thebibliography}{8}
% % \bibitem{ref_article1}
% % Author, F.: Article title. Journal \textbf{2}(5), 99--110 (2016)

% % \bibitem{ref_lncs1}
% % Author, F., Author, S.: Title of a proceedings paper. In: Editor,
% % F., Editor, S. (eds.) CONFERENCE 2016, LNCS, vol. 9999, pp. 1--13.
% % Springer, Heidelberg (2016). \doi{10.10007/1234567890}

% % \bibitem{ref_book1}
% % Author, F., Author, S., Author, T.: Book title. 2nd edn. Publisher,
% % Location (1999)

% % \bibitem{ref_proc1}
% % Author, A.-B.: Contribution title. In: 9th International Proceedings
% % on Proceedings, pp. 1--2. Publisher, Location (2010)

% % \bibitem{ref_url1}
% % LNCS Homepage, \url{http://www.springer.com/lncs}, last accessed 2023/10/25
% % \end{thebibliography}

\end{document}